\documentclass[letterpaper, 10 pt, conference]{ieeeconf}  

\IEEEoverridecommandlockouts                              

\title{\LARGE \bf Contact-GraspNet: Efficient 6-DoF Grasp Generation \\ in Cluttered Scenes
}

\author{Martin Sundermeyer$^{1,2,3}\qquad$ Arsalan Mousavian$^1\qquad$ Rudolph Triebel$^{2,3}\qquad$ Dieter Fox$^{1,4}$ 

\thanks{$^*$This work is done while the first author was an intern at NVIDIA.
$^1$NVIDIA {\tt\small (amousavian,dieterf)@nvidia.com}, $^2$German Aerospace Center (DLR) {\tt\small <first>.<last>@dlr.de}, $^3$Technical University of Munich (TUM), $^4$University of Washington
}}
\usepackage{hyperref}

\usepackage{cite}
\usepackage{breakcites}
\usepackage{graphicx}
\usepackage{amssymb}
\usepackage{amsmath}
\usepackage{amsfonts}
\usepackage{booktabs}
\usepackage{siunitx}
\usepackage{multirow}

\usepackage[utf8]{inputenc}
\usepackage{pgfplots}
\DeclareUnicodeCharacter{2212}{−}
\usepgfplotslibrary{groupplots,dateplot}
\usetikzlibrary{patterns,shapes.arrows}
\pgfplotsset{compat=newest}
\tikzset{every picture/.style={/utils/exec={\sffamily}}}
\tikzset{font=\footnotesize}

\DeclareMathOperator*{\argmin}{arg\,min}
\let\vec\mathbf

\newlength\figH
\newlength\figW
\begin{document}

\maketitle
\thispagestyle{empty}
\pagestyle{empty}

\begin{abstract}

Grasping unseen objects in unconstrained, cluttered environments is an essential skill for autonomous robotic manipulation.
Despite recent progress in full 6-DoF grasp learning, existing approaches often consist of complex sequential pipelines that possess several potential failure points and run-times unsuitable for closed-loop grasping.
Therefore, we propose an end-to-end network that efficiently generates a distribution of 6-DoF parallel-jaw grasps directly from a depth recording of a scene.
Our novel grasp representation treats 3D points of the recorded point cloud as potential grasp contacts. 
By rooting the full 6-DoF grasp pose and width in the observed point cloud, we can reduce the dimensionality of our grasp representation to 4-DoF which greatly facilitates the learning process.
Our class-agnostic approach is trained on 17 million simulated grasps and generalizes well to real world sensor data. In a robotic grasping study of unseen objects in structured clutter we achieve over 90\% success rate, cutting the failure rate in half compared to a recent state-of-the-art method. Video of the real world experiments and code are available at~\url{https://research.nvidia.com/publication/2021-03_Contact-GraspNet\%3A--Efficient}.

\end{abstract}


\section{Introduction}

The ability to grasp objects is one of the fundamental capabilities required in most robot manipulation tasks. Grasping involves reasoning about the 3D geometry and physics properties of the object such as mass and friction, and also reasoning about complex contact physics. It is studied in two main directions: Model-based grasping where the 3D model or category of the object is known and model-free grasping where there is no prior knowledge about the object. Model-based grasping circumvents reasoning about the physics of contact and grasp generation by pre-defining a set of grasps in the object frame and transform those grasps according to the 6-DoF object pose \cite{deng2019pose, sundermeyer2020augmented, deng2019self,wang2019densefusion} or detected keypoints of the objects \cite{manuelli2019kpam,fang2018tog}. The downside of model-based approaches is that they only work on a limited subset of known objects or categories, and any errors in detecting 6-DoF object pose or object keypoints degrade the grasping performance.

 \begin{figure}[t]
    \centering
    \includegraphics[width=0.45\textwidth, angle=0]{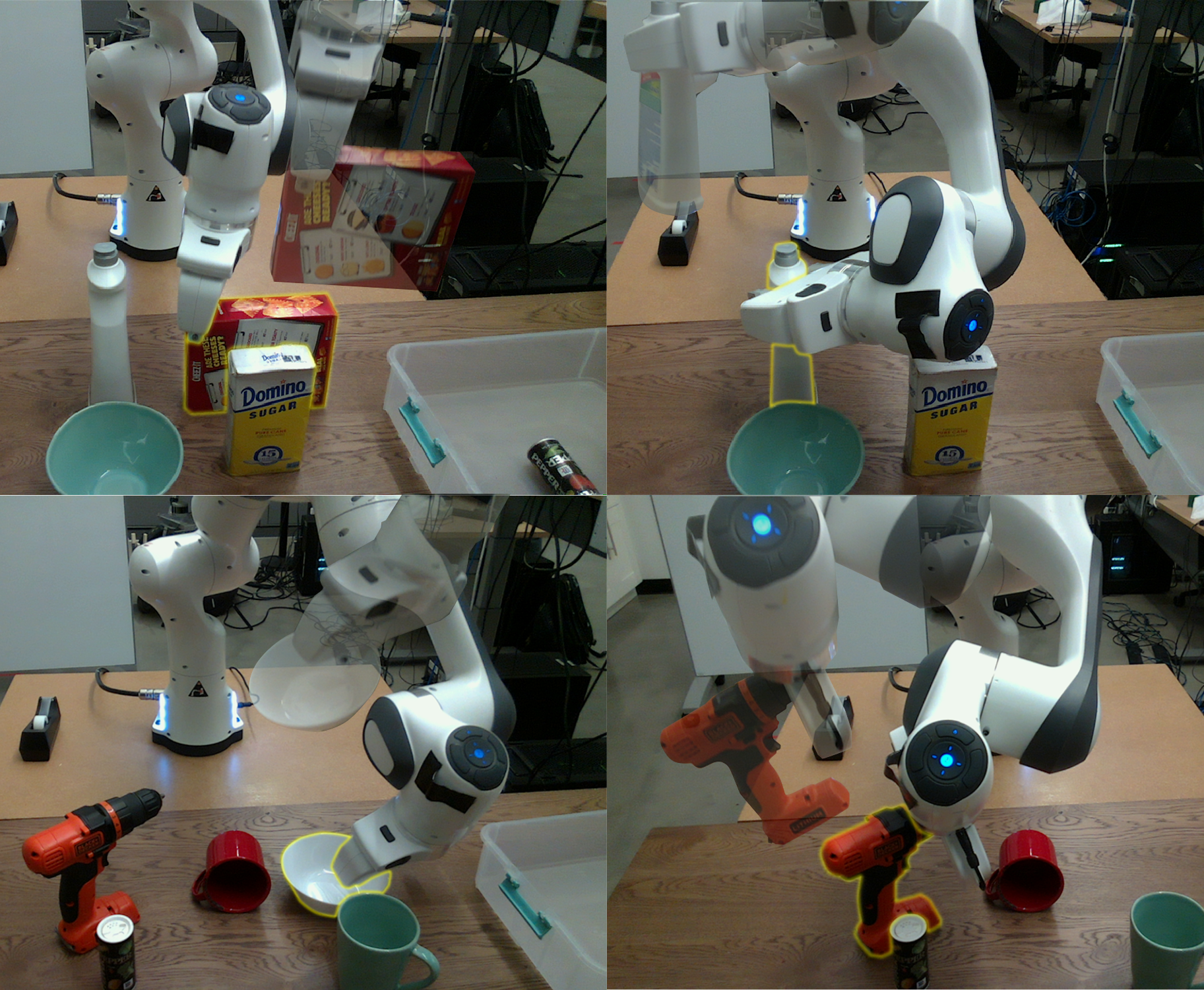}
    \caption{Contact-GraspNet efficiently predicts diverse and stable grasps in cluttered scenes while avoiding collisions.}
    \label{fig:cover}
\end{figure}

Model-free approaches do not make any strong assumptions about the category or shape of the object, and they learn a shared representation for all object shapes and sizes. However, having one shared representation for all objects in addition to the large SE(3) space for the grasp poses makes the learning problem quite challenging. As a result, a large body of work in data-driven grasping constraints the space of possible grasps to planar grasping, where grasps are represented by oriented rectangles around each pixel that define the grasp frame \cite{mahler2017dex,mahler2019learning, lenz2015deep}. Such a representation needs the camera to view the scene perpendicularly and thus limits 3D reasoning and applications significantly. A large number of possible grasps and the full kinematic capabilities of the robot are also neglected. To address the limitations of planar grasping, there has been a recent interest in  tackling the problem of 6-DoF grasping of unknown objects \cite{ten2017grasp,mousavian20196,murali20206,song2020grasping,fang2020graspnet}. In this paper, we tackle 6-DoF grasping of unknown objects in cluttered space from a partial point cloud observation of the scene.

Grasping objects from cluttered scenes with structure introduces extra challenges. The target objects must be grasped successfully, while at the same time any collision with other objects must be avoided to prevent damages or transformations into other undesired states. This is particularly important in home robotics and healthcare applications. Additionally, it is crucial to generate a diverse set of grasps for the object due to robot kinematic constraints. Depending on the relative pose between the object and the robot, a different subset of grasps is kinematically feasible. 

Our method is closely related to the work of Murali \emph{et al.}~\cite{murali20206}, where the goal is to generate collision-free diverse grasps for a designated target object from a partial point cloud of the scene, and the objects are segmented using a pre-trained unknown object instance segmentation model \cite{Xiang2020LearningRF,XieCoRL19}. Murali \emph{et al.}~\cite{murali20206} use a multi-stage process that synthesizes grasps for the target objects from the segmented object point cloud with no context around it, and then filters out the colliding grasps using another learned model. This leads to three issues: 1) Sensitivity to instance segmentation errors. 2) Grasps are generated just from the target object point cloud and do not leverage geometric cues in the scene such as table points and surrounding object points. 3) Grasps are predicted in the large, unconstrained 6-DoF pose space.
To address these issues, our method instead directly processes a full scene point cloud or a local region around a target object. Therefore, the quality of our generated grasps is not depending on an accurate mask and collisions can be directly taken into account during generation. Instance segmentation can then subsequently be used to filter grasps belonging to a target object.  Thus, our main contributions are the following:
\begin{itemize}
    \item A new end-to-end method for 6-DoF grasping of unknown objects in cluttered real world scenes where we achieve $90\%$ grasp success rate. This is $10\%$ higher than \cite{murali20206} in equal settings.
    \item A new grasp pose representation that projects 6-DoF grasps to their contact points in an observed point cloud. Our representation has only 4-DoF which facilitates the learning problem significantly.
    \item Comprehensive ablation studies in a physics simulator to evaluate the effects of different loss functions and training data.
\end{itemize}

\section{Related Work}
As a fundamental problem in robotics, grasping has been studied for decades \cite{inbook, 844081, 6942775, du2019vision}. We review related literature in the context of data-driven methods.

\noindent {\bf End-to-end policy learning:}
One line of work for grasping and manipulation of objects employs an end-to-end policy that learns to generate actions from raw input pixel values \cite{levine2016learning, qtopt2018}. This results in a monolithic model that concurrently reasons about perception, planning, grasping, and controlling the robot. A large group of these works learn from interactions of the robot with the environment through reinforcement learning. These approaches have mostly shown promise in bin picking, in (quasi-) planar grasping and in small, insensible workspaces that do not require complicated motion planning in the robot configuration space. Few works \cite{zeng2018learning} have demonstrated iterative 6-DoF grasping approaches with a monolithic policy by combining imitation learning and reinforcement learning. A common drawback of these methods is the limited generalization to novel environments, because the perception and control are learned indirectly at the same time. In addition, these methods are not easily steerable towards grasping a specific object as the reward function encourages grasping any object. 
In contrast, our method learns to generate diverse 6-DoF grasps on novel objects and scenes for specifiable target objects while just using simulated training data. Additionally, it can be integrated with other perception and motion planning algorithms.

\noindent {\bf 3D reconstruction:}
A complete 3D reconstruction enables traditional grasp planning. However, learned single-view reconstructions are often ambiguous, coarse and require class-conditioning \cite{yan2019dataefficient, yan2018learning, agnew2020amodal}. Multiple views for 3D scanning are beneficial \cite{breyer2021volumetric} but not always obtainable, take additional time and typically assume a static scene. In our approach a full explicit 3D reconstruction is not required. 

\noindent {\bf Discriminative methods:} Discriminative methods for grasping train a classifier that evaluates the quality of existing grasps \cite{doi:10.1177/0278364915577105, liang2019pointnetgpd, mahler2017dex}. They use different sampling strategies to generate potential candidates. For planar grasping, cross entropy is widely used since it can converge to the final grasp location by iteratively evaluating the quality of grasps in different locations~\cite{mahler2017dex}. However, the cross-entropy method does not work well in the higher dimensional 6-DoF grasp space. To overcome the sampling complexity issue, grasp locations are often sampled using geometric heuristics~\cite{ten2017grasp,liang2019pointnetgpd}. 

\noindent {\bf Generative methods:} Learning-based generative grasp methods aim to overcome the limitations of geometric heuristics and generate meaningful 6-DoF grasps often from experience in a physics simulator~\cite{mousavian20196,murali20206}. The main challenge is the large, multi-modal search space of 6-DoF grasps. Instead of sampling some potential candidates using heuristics and ranking them, these models directly predict a per-point graspability score and approach direction in SO(3) space ~\cite{qin2020s4g, fang2020graspnet, ni2020pointnet++}. One problem with predicting approach directions is that they cannot easily capture high curvature areas such as mug rims or handles and also can not represent grasps encompassing hollow structures. Furthermore, successful approach directions are quite ambiguous to learn as multiple ones are possible for a single contact. Instead we propose to predict 6-DoF grasps densely projected to their much less ambiguous contact points. While grasps without full surface contact are plausible, e.g. through the handle of a mug, the knowledge about the object state and therefore the ability to steadily place the object again is lost. Therefore, in this work we are aiming to generate stable grasps for unknown objects with full surface contact. 
Our novel loss formulation further improves convergence by accounting for the discontinuities, imbalance and multi modality of the grasp distribution. Unlike other methods \cite{ni2020pointnet++}, our proposed method is independent of category labels and has no assumption of grasps being always perpendicular to a surface. Instead we learn a grasp semantic purely from a wide variety of grasp annotated training shapes \cite{eppner2021icra}.





\section{Method}
\begin{figure*}[t]
    \centering
    \includegraphics[width=1.0\textwidth, angle=0]{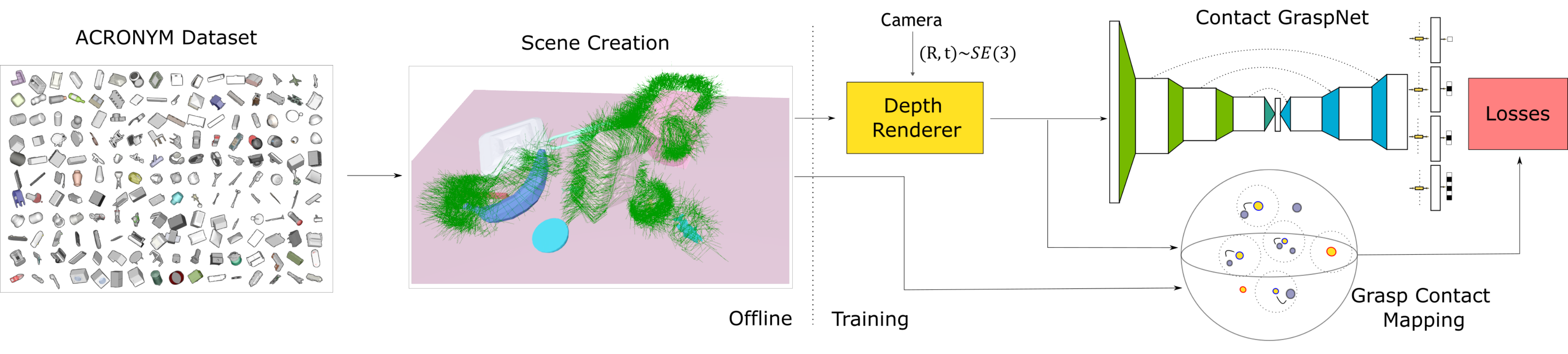}
    \caption{Training Data Pipeline. We place object meshes with dense grasp annotations from the ACRONYM dataset \cite{eppner2021icra} at random stable poses in scenes. Grasp poses that produce gripper model collisions are removed. Resulting grasps are mapped to their contacts on the mesh surface. During training, we sample virtual cameras to render point clouds from the scenes. We consider recorded points (yellow) as positive contacts if there exists a mesh contact (blue) in a 5mm radius and associate the grasp transformation belonging to the closest mesh contact to them. These per-point annotations are used to supervise the Contact Grasp Network.}
    \label{fig:data_pipeline}
\end{figure*}%

 \begin{figure}[t]
    \centering
    \includegraphics[width=0.4\textwidth, angle=0]{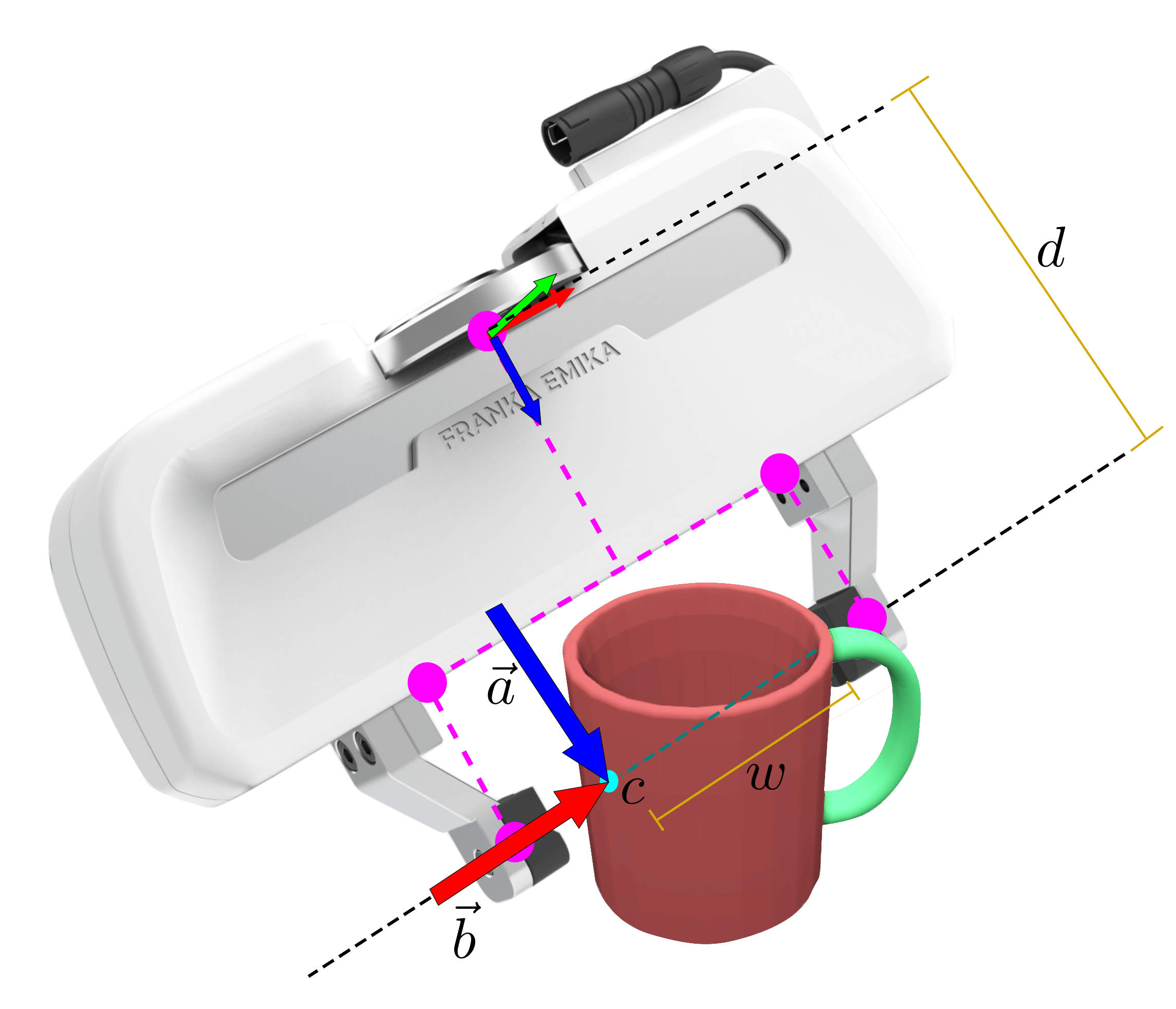}
    \caption{Our grasp representation: $c$ depicts an observed contact point. $\vec{a}$ and $\vec{b}$ constitute the 3-DoF rotation, $w$ is the predicted grasp width, $d$ the distance from baseline to base frame. In pink we show the five gripper points $\vec{v}$ that we used in the $l_{add-s}$ loss.}
    \label{fig:grasp_representation}
\end{figure}

We consider the problem of generating 6-DoF grasps from any viewpoint on structured clutter consisting of unknown objects. Our approach takes in a raw depth image, optionally with object masks, and generates 6-DoF grasp proposals together with corresponding grasp widths. Our goal is to predict grasps that are robust, diverse and non-colliding from an only partially observable scene. 

\textbf{From a learning perspective}, generating the distribution of successful 6-DoF grasps is quite challenging, because the distribution is multi-modal, discontinuous, imbalanced and ambiguous due to (self-) occlusions. Furthermore, direct regression in high dimensional output spaces like $SE(3)$ has been shown to be difficult in grasping \cite{mousavian20196} and also in related fields such as object pose estimation \cite{sundermeyer2018implicit}. 

\subsection{Grasp Representation}

For these reasons, finding an efficient grasp representation is crucial to solve this task using learning-based methods. This representation should generalize well to unseen objects and handle the high-dimensional output space well.

%

\textbf{Contact Grasp Representation: }
We observe that for most predictable two-finger grasps at least one of the two contacts is visible prior to grasping. In contrast, grasps without any visible contact are often ambiguous or do not preserve the initial object pose after grasping. 
Therefore, we map a distribution of successful 6-DoF ground truth grasps $g \in G$ to their corresponding contact points $c \in \mathbb{R}^3$.
Since visible contact points are bound to lie on surfaces that we can observe with a depth sensor, we can represent their 3D location by nearby points in a recorded point cloud. 

Given that we can predict whether observed points are suitable grasp contacts, we can thus reduce the 6-DoF grasp learning problem to estimating the 3-DoF grasp rotation $R_{g} \in \mathbb{R}^{3 \times 3}$ and grasp width $w \in \mathbb{R}$ of a parallel-yaw gripper.

Starting from a contact point $\vec{c} \in \mathbb{R}^3$, where the gripper baseline intersects the mesh, we depict a 6-DoF grasp pose $g \in G$ defined by $(R_g,t_g) \in SE(3)$ and grasp width $w \in \mathbb{R}$ as

\begin{equation}
    \label{eq:t}
    \vec{t}_{g} = \vec{c} + \frac{w}{2}\vec{b} + d \vec{a}
\end{equation}
\begin{equation}
    \label{eq:R}
    R_{g} = \begin{bmatrix}
    \vert & \vert & \vert\\
    \vec{b}   &  \vec{a} \times \vec{b}  &  \vec{a} \\
    \vert & \vert & \vert \\
\end{bmatrix},
\end{equation}
where $\vec{a} \in \mathbb{R}^3, ||\vec{a}||=1$ is the approach vector, $\vec{b} \in \mathbb{R}^3, ||\vec{b}||=1$ is the grasp baseline vector, and $d \in \mathbb{R}$ is the constant distance from the gripper baseline to the gripper base. Our grasp representation is depicted in Figure \ref{fig:grasp_representation}.

 The reduced dimensionality greatly facilitates the learning process compared to methods that estimate grasp poses in unconstrained $SE(3)$ space. It also increases the pose accuracy of predicted grasps as they are bound to the geometry of the observed scene. 
 In contrast to axis-angle representations, our rotation representation has neither ambiguities nor discontinuities.
  Moreover, at test time we can sample grasp proposals by sampling contact points that cover the whole observable surface of the scene/object and thus represent the modes of the 6-DoF grasp distribution well. While a 3D view on the scene is preferable, even a frontal view on a box produces reasonable grasps due to the radial mapping.
 
\textbf{Point Set Networks}
such as PointNet++ \cite{qi2017pointnetplusplus} effectively process point clouds and hierarchically aggregate points and their feature representations in local 3D neighborhoods. Their predictions can be directly associated to 3D points in the input point cloud and our proposed grasp representation exploits this ability.

\subsection{Data Generation}

To learn the full distribution of stable 6-DoF grasps, diverse and dense grasp pose annotations are required. We used the ACRONYM dataset~\cite{eppner2021icra}, which consists of 8872 meshes from the Shapenet dataset~\cite{chang2015shapenet} and $17.7$ million simulated grasps under varying friction.
An overview of our offline and online training data generation is given in Fig. \ref{fig:data_pipeline}. 


During training we render a scene point cloud $\mathcal{P}=\{\vec{p}_1,\dots,\vec{p}_n\} \subset \mathbb{R}^{3}$ and assign a point-wise grasp success
\begin{equation}
    \forall i=1,\dots, n \quad s_i = \begin{cases}1 & \min_j ||\vec{p}_i-\vec{c}_j||_2 < r \\ 
                          0 & otherwise, \end{cases} 
\end{equation}
where $\vec{c}_j \in \mathcal{P}$ 
are the mesh contact points of non-colliding ground truth grasps $g_j \in G$ in camera coordinates and  $r \in \mathbb{R}$ is their maximum propagation radius. Thus, $\mathcal{P}$ can be split into points $\mathcal{P}^-:=\{\vec{p}_i|s_i = 0\} $, where no feasible grasp contact is found within a radius of $r=5mm$, and $\mathcal{P}^+:=\{\vec{p}_i|s_i = 1\}$, containing points suitable for a contact. To the latter ones $\vec{p}^+_i\in\mathcal{P}^+$ we assign the closest grasp as

  \begin{align}
  \label{eq:closest_grasp}
    \begin{bmatrix}
           w_{g,i} \\
           R_{g,i} \\
           \vec{t}_{g,i} 
         \end{bmatrix} &= 
         \begin{bmatrix}
           w_{g,j} \\
           R_{g,j} \\
            \vec{p}^+_{i} + \frac{w_j}{2}\vec{b}_j + d \vec{a}_j 
         \end{bmatrix}
  \end{align}
with 
 \begin{align}
j=\argmin_k || \, \vec{p}^+_{i}-\vec{c}_k||_2
\end{align}


Given sufficient coverage we can thereby project the ground truth distribution of 6-DoF grasps densely on the recorded point cloud.

\subsection{Network}
We employ the set abstraction and feature propagation layers proposed in PointNet++ \cite{qi2017pointnetplusplus} to build an asymmetric U-shaped network. The network takes n=20000 random points $p \in \mathbb{R}^{20000 \times 3}$ as input and predicts grasps for only m=2048 farthest points of the input to make sure the inference fits in GPU memory and predicted grasps have good coverage over the scene. The network has four heads with two 1D-Conv layers each and per-point outputs $s \in  \mathbb{R}, \vec{z}_1 \in \mathbb{R}^3, \vec{z}_2 \in \mathbb{R}^3, \vec{o} \in \mathbb{R}^{10}$, from which we form our grasp representation.
%
%
%
%
The predicted grasp width $\vec{\hat{w}}_i \in [0,w_{max}]$ is split into 10 equidistant grasp width bins $\vec{\hat{o}} \in \mathbb{R}^{10}$ to counteract data imbalance. Then, $\vec{\hat{w}}_i$ is represented by the center value of the bin(s) with the highest confidence.
The approach direction $\vec{a} \in \mathbb{R}^3$ and the baseline direction $\vec{b} \in \mathbb{R}^3$ are orthonormal by definition. We inject this property into training by coupling the predictions $\vec{\hat{a}}, \vec{\hat{b}}$ through an in-network Gram Schmidt orthonormalization

\begin{align}
    \vec{\hat{b}} = & \frac{\vec{z}_1}{||\vec{z}_1||} &
    \vec{\hat{a}} = \frac{\vec{\vec{z}_2 - \langle \hat{b}, \vec{z}_2 \rangle \hat{b}}}{||\vec{z}_2||}
\end{align}
Thus, we perform a projection and only predict $\vec{\hat{a}}$ as the component that is orthonormal to $\vec{\hat{b}}$. The orthonormalization further reduces the dimensionality of our predicted grasp representation and facilitates the regression of 3D rotations \cite{zhou2019continuity}.

 \begin{figure*}[t]
    \centering
    \includegraphics[width=1.\textwidth, angle=0]{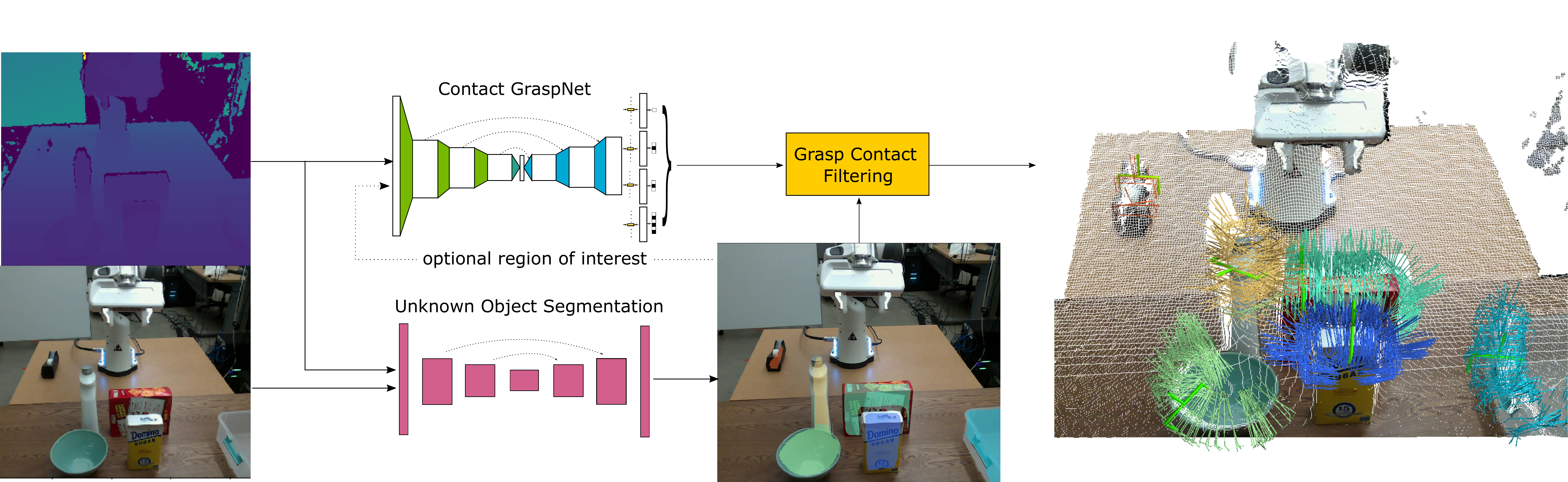}
    \caption{Full Inference Pipeline: We segment unknown objects from an RGB-D image using \cite{Xiang2020LearningRF}. Our Contact-GraspNet processes the full scene point cloud or a local region of interest around a target object. Predicted 6-DoF grasps are then associated to object segments by filtering their contact points. On the right we show the predicted 6-DoF grasp distribution and, in bold, the most confident grasp per segment.\vspace{-5mm}}
    \label{fig:inference}
\end{figure*}

\begin{figure}
\begin{center}
\scalebox{0.89}{
\setlength{\figH}{6.5cm} 
\input{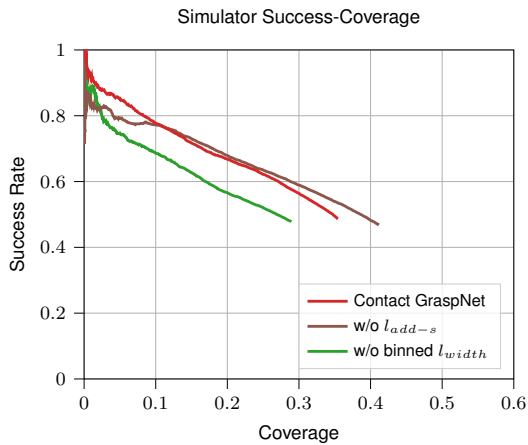}}
\end{center}
\vspace{-4mm}
\caption{Loss Ablations: Without weighted binning in the grasp width loss $l_{width}$ both, success rate and coverage decrease. The $l_{add-s}$ loss leads to increased success rates at high confidence contacts (Coverage $\in [0,0.1]$) and to slightly decreased success rate in the low-confidence regime. This confidence calibration is important, since it determines which grasp is eventually executed.}
\label{fig:loss_ablation}
\end{figure}

\begin{figure}
\vspace{-10mm}
\begin{center}
\scalebox{0.89}{
\setlength{\figH}{6.5cm} 
\input{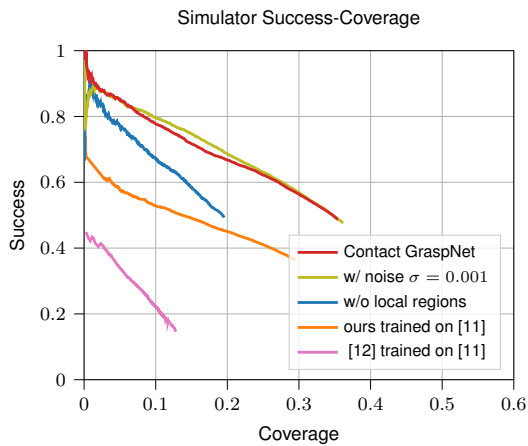}}
\end{center}
\vspace{-4mm} 
\caption{Data Ablations: Training with Gaussian noise has similar performance in simulation but helps generalization to noisy sensor data. Predicting grasps directly on full scenes without extracting local regions yields a similar average success rate, but significantly lowers grasp coverage. Training on the small grasp dataset from \cite{mousavian20196} with 5 categories is not sufficient to generalize to arbitrary objects and shows the importance of ACRONYM \cite{eppner2021icra} \vspace{-6mm}}
\label{fig:data_ablation}
\end{figure}

\subsection{Target Losses}

The contact grasp success predictions $\hat{s} \in \mathbb{R}$ are evaluated at all output points $ \vec{p}_i \in \mathbb{R}^3 : \forall i \in [0,m]$ using binary cross entropy. We only backpropagate the top-k point predictions with the largest errors $l_{bce,k}$, with k=512, to counteract data imbalance. The other predictions concerning the geometry of grasps are only evaluated at positive contact points $\vec{p}^+_i$.  
Instead of supervising all network heads in isolation, we propose to combine the predictions to the 6-DoF grasp pose $\hat{g} \in G$ given in Eq. \eqref{eq:t} and \eqref{eq:R} already during training. We define five 3D points $\vec{v} \in \mathbb{R}^{5 \times 3}$ representing the 6-DoF gripper pose, as shown in Fig. \ref{fig:grasp_representation}, and transform these using all ground truth and predicted grasp poses defined in Eq. \eqref{eq:closest_grasp}
\vspace{-8mm}

\begin{align}
    \vec{v}^{gt}_i = & \vec{v} R_{g,i}^T + \vec{t}_{g,i}& 
    \vec{v}^{pred}_i = \vec{v} \hat{R}_{g,i}^T + \vec{\hat{t}}_{g,i} 
\end{align}

We formulate the 6-DoF grasp loss $l_{add-s}$ as a weighted minimum average distance between gripper points $\vec{v}^{gt}$ and $\vec{v}^{pred}$ where we take the symmetry of the gripper into account.
\vspace{-4mm}

\begin{align}
    l_{add-s} = \frac{1}{n^+} \sum^{n^+}_i \hat{s_i} \min_u || \vec{v}^{pred}_i - \vec{v}^{gt}_u ||_2,
\end{align}
where
$n^+$ is the size of $\mathcal{P}^+$. We weight each distance to the closest ground truth grasp points with the predicted contact success confidence $\hat{s}_i$. 

Our proposed loss formulation has several advantages: (1) We can learn the different modes of the ground truth grasp distribution, e.g. different predicted grasp approach directions $\hat{\vec{a}}$ can produce a small error. (2) The point-wise weighting with $\hat{s}_i$ couples the contact point classification with the grasp pose predictions. Contact confidence can only increase if the network predicts a 6-DoF grasp pose close to a ground truth pose. (3) Wrongly predicted grasps in regions far away from any ground truth grasp, e.g. at artificial edges from occlusions, produce a high loss and are thus avoided. 

On the grasp width bin predictions, we optimize a weighted, multi-label binary cross entropy loss $l_{width}$. Since small grasp widths are highly over-represented, we weight the bin losses anti-proportional to bin size. Our total loss is $l = \alpha l_{bce,k} + \beta l_{add-s} + \gamma l_{width}$ with $\alpha=1, \beta=10, \gamma=1$.



    


\subsection{Implementation Details}

We use the Adam optimizer with an initial learning rate of 0.001 and a step-wise decay to 0.0001. Our set abstraction layers have 3 parallel branches with query ball radii [0.02,0.04,0.08], [0.04,0.08.0.16] and [0.08,0.16,0.32]. For inference the point cloud is centered at its mean in camera coordinates.
For training we generate 10000 table top scenes by placing 8-12 grasp annotated ShapeNet models \cite{eppner2021icra} at random stable poses. We use rejection sampling to avoid collisions. 
We train with a batch size of 3 for 144.000 iterations which takes $\sim40$ hours on a single Nvidia V100 GPU. Convergence is significantly faster than on previous methods~\cite{murali20206,fang2020graspnet,mousavian20196} which take up to one week on a single GPU for training. This also reflects the effectiveness of our proposed grasp representation.

\section{Experimental Evaluation}

We evaluate our method in a grasping study with a Franka robot where we pick unknown objects in cluttered scenes. We also compare different variations of our method and of our data by executing a large number of predicted grasps in the FleX physics simulator \cite{macklin2014unified}. 

\subsection{Inference}
Our inference pipeline is shown and described in Fig. \ref{fig:inference}. The Contact-GraspNet can also be applied to raw depth images by itself, but most robotic tasks require some kind of instance detection/segmentation to specify a target.

\textbf{Local regions} of interest can be optionally extracted around the 3D centroid of point cloud segments in order to maximize the number of potential contact points. In our experiments, we extract cubes with an edge length set to twice the largest spanning dimension, but at least $0.3m$ and at most $0.6m$.

\textbf{Run time:} The Contact-GraspNet has a run time of $0.28s$ for a full scene or $\sim 0.19s$ for a local region around a target object. Compared to other 6-DoF grasp generation methods this is quite fast and enables applications requiring reactive closed loop grasping.

\textbf{Grasp Selection:} At test time we select grasps by setting a contact confidence threshold of $0.23$ and then use farthest point sampling on the (filtered) contact points to ensure broad grasp coverage. If the number of predicted grasps for an object is too low, we reduce the confidence threshold to $0.19$. In the end we execute the most confident grasp that is kinematically reachable and where the robot does not collide with the scene \cite{danielczuk2021icra}.

 \begin{figure}[t]
    \centering
    \includegraphics[width=0.4\textwidth, angle=0]{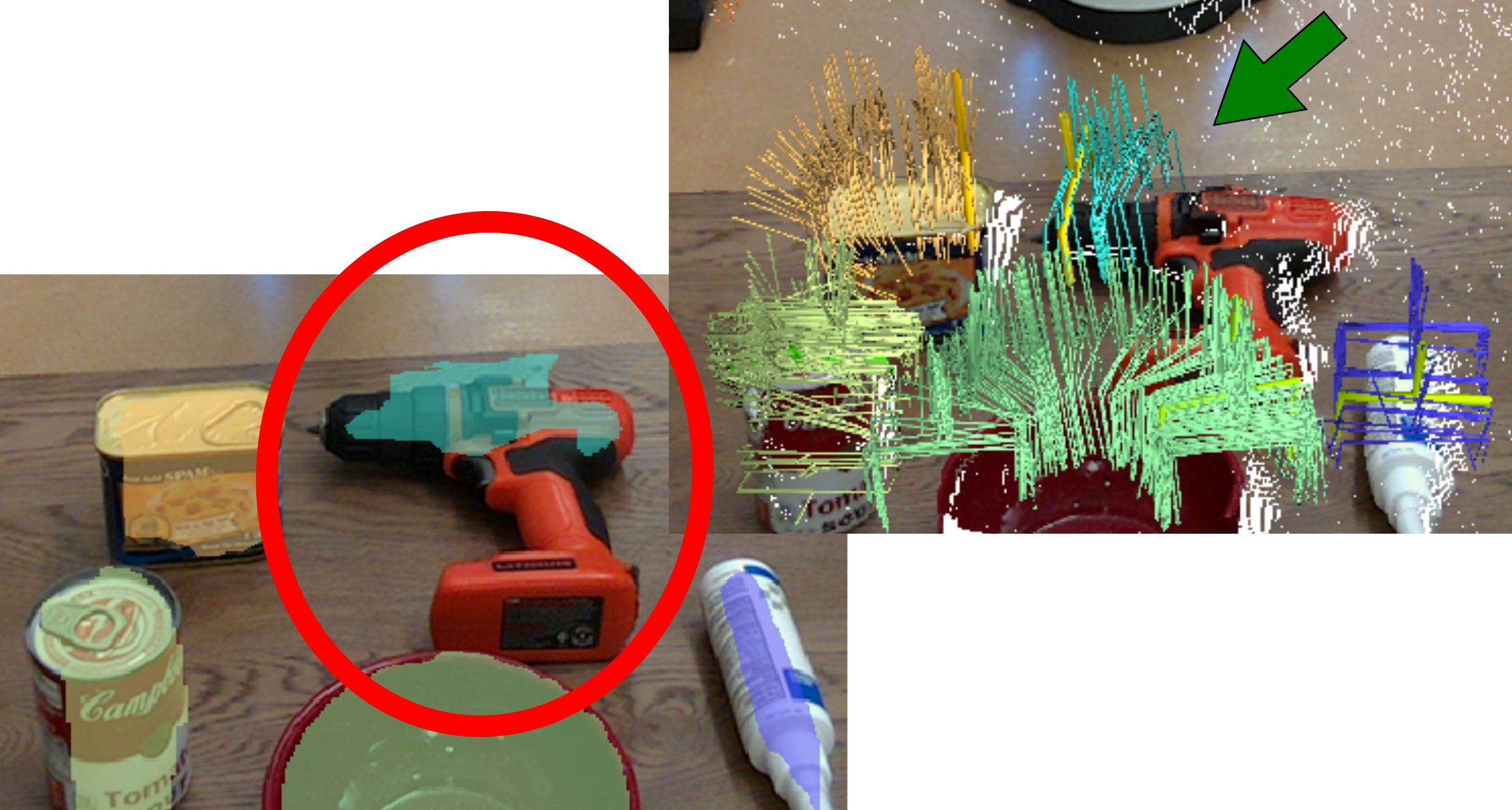}
    \caption{One advantage of our method is that it does not rely on an accurate segmentation of unknown objects. Here, successful grasp contacts are still found on the driller despite severe under-segmentation.}
    \label{fig:incomplete_segment}
\end{figure}

\subsection{Evaluation Metrics}
\textbf{In our robotic experiments} we report the number of successful grasps and the number of trials. The latter is often disregarded when picking small objects from a bin. However, grasping in only one or two trials is crucial in cluttered scenes (e.g. in households) with large, densely packed objects where collisions should be avoided and stable grasp opportunities can vanish after objects tip over. We limit ourselves to a maximum of two grasp trials per object without rearrangements and report the success rate after a single trial as well.

\textbf{Our simulator experiments} allow us to also evaluate the diversity of grasps and ablate variations of our method. Here, we evaluate the success rate and coverage of the generated grasps following \cite{mousavian20196}. A grasp is considered successful if (1) the open gripper does not collide with the object/scene and (2) the object is still in the gripper after grasping and a shaking motion. This is a conservative measure, as most real world grasps can slightly collide and do not undergo a shaking motion. Coverage is the percentage of ground truth grasps (including occluded ones) whose base coordinates are within 2cm of any of the generated grasps. 

\subsection{Real robot grasp experiments}

\textbf{Setup:} Our physical setup consists of a 7-DoF Franka Panda robot with a parallel-jaw gripper. We closely replicate the 9 cluttered scenes defined in \cite{murali20206} with a total of 51 unseen objects. The task is to pick the objects from the cluttered scene and place them into a bin. We manually select target objects and grasp them in the same random order as in \cite{murali20206}. In our experiments, we use the Intel Realsense L515 LiDAR camera mounted on a tripod for both RGB and depth data. Robot motions are generated using~\cite{danielczuk2021icra}.

\textbf{Results:} Table \ref{tab:results} shows our grasp evaluation results on the robot. We observe a significantly higher grasp success rate of our method compared to \cite{mousavian20196} and \cite{murali20206} which themselves outperform other learning-based methods and analytic/heuristic baselines. Furthermore, our method strongly improves the grasp success at first trial and thereby reduces the number of re-grasps. 
 We also addressed the shortcomings of cropping objects from the point cloud using potentially imprecise segmentation masks. Fig. \ref{fig:incomplete_segment} shows an imprecise segmentation example where cropping would be catastrophic but where our grasp filtering method can still extract successful grasps. 
 
\subsection{Ablations}
\textbf{Optimization Targets:} In Fig. \ref{fig:loss_ablation} we first investigate the effect of our loss targets. The weighted loss on the grasp width bins $l_{width}$ is crucial to deal with the imbalanced widths in our grasp dataset. Without weighting the bins, the predictions mostly collapse into narrow grasp widths. Weighting also performs better than oversampling in our experiments. The average distance loss $l_{add-s}$ improves the success rate of high confidence contacts which is important because most grasps that we execute lie in the first decimal of coverage. The connection of contact confidence with the grasp pose results in an overall improved calibration.

\textbf{Data:} In Fig. \ref{fig:data_ablation} we examine the effects of different training and test data. Zooming into local regions allows the network to concentrate potential contact points on the object and thus increases coverage. We also show the importance of a large and diverse grasp dataset like ACRONYM \cite{eppner2021icra}. Training on a small grasp datasets with $110$ objects from $5$ categories \cite{mousavian20196} is not sufficient for out-of-category generalization irrespective of the method.

\textbf{Failure Cases:} We observe some failure cases for thick objects that only allow grasps almost at maximum grasp width. Here, grasp predictions are less confident presumably because of the discontinuous decision boundary. Injecting noise during training reduces this effect. Finally, small objects sometimes have contact points with low confidence possibly because of their small impact on the total loss.
\begin{table}[t]
\caption{Cluttered Scene Grasping: We achieve a clear improvement over recent state-of-the-art grasping pipelines}
\small
\begin{tabular}{ccccc} \toprule
    & Success & First attempt & \#Attempts \\ \midrule
     
        6-DOF GraspNet\cite{mousavian20196} & 62.7 & - & - \\
    \multirow{2}{*}{\cite{mousavian20196} +CollisionNet\cite{murali20206}} & \multirow{2}{*}{80.39} & \multirow{2}{*}{68.63} & \multirow{2}{*}{67}  \\ 
 &&& \\
    Contact-GraspNet & $\boldsymbol{90.20}$ & $\boldsymbol{84.31}$ & $\boldsymbol{59}$ \\
     \bottomrule
\end{tabular}
\label{tab:results}
\end{table}
\section{Conclusions}

We considered the fundamental problem of grasping unknown objects in structured clutter with a parallel jaw gripper.
We proposed an efficient, accurate and simplifying 6-DoF grasp generation method called Contact-GraspNet. By transforming the hardly tractable 6-DoF grasp estimation problem into a grasp contact point classification and a grasp rotation estimate, we greatly limit the predicted pose space and facilitate the learning process. Through tailored optimization targets that take into account the multi-modality, imbalance and sparsity of the 6-DoF grasp distribution, our network learns to generate diverse grasps covering the whole graspable surface in a recorded scene. Gripper collisions are effectively avoided by considering them during training and by predicting grasps directly in scenes. Our approach can incorporate segmentation predictions as well but is not dependent on accurate masks itself. It is also complementary to grasp ranking methods that use gripper and/or robot models as input.
Grasping successfully with a single attempt is crucial in sensible environments. Our method showed strong advances in that regard and is a step towards reaching the required grasp reliability.  


\bibliographystyle{IEEEtran}
\bibliography{IEEEabrv,references}

\end{document}